# Automated Transcription for Pre-Modern Japanese Kuzushiji Documents by Random Lines Erasure and Curriculum Learning

Anh Duc Le[1]

[1] Center for Open Data in The Humanities, Tokyo, Japan
`anh@ism.ac.jp`

**Abstract.** Recognizing the full-page of Japanese historical documents is a challenging problem due to the complex layout/background and difficulty of writing styles, such as cursive and connected characters. Most of the previous methods divided the recognition process into character segmentation and recognition. However, those methods provide only character bounding boxes and classes without text transcription. In this paper, we enlarge our previous human-inspired recognition system from multiple lines to the full-page of Kuzushiji documents. The human-inspired recognition system simulates human eye movement during the reading process. For the lack of training data, we propose a random text line erasure approach that randomly erases text lines and distorts documents. For the convergence problem of the recognition system for full-page documents, we employ curriculum learning that trains the recognition system step by step from the easy level (several text lines of documents) to the difficult level (full-page documents). We tested the step training approach and random text line erasure approach on the dataset of the Kuzushiji recognition competition on Kaggle. The results of the experiments demonstrate the effectiveness of our proposed approaches. These results are competitive with other participants of the Kuzushiji recognition competition.

**Keywords:** Kuzushiji recognition, full-page document recognition, random text line erasure approach.

## 1 Introduction

Japan had been using Kuzushiji or cursive writing style shortly after Chinese characters got into the country in the 8th century. Kuzushiji writing system is constructed from three types of characters which are Kanji (Chinese character in the Japanese language), Hentaigana (Hiragana), and Katakana, like the current Japanese writing system. One characteristic of classical Japanese, which is very different from the modern one, is that Hentaigana has more than one form of writing. For simplifying and unifying the writing system, the Japanese government standardized Japanese language textbooks in 1900 [1]. This makes the Kuzushiji writing system is incompatible with modern printing systems. Therefore, most Japanese natives cannot read books written by Kuzushiji. We have a lot of digitized documents collected in libraries and museums throughout the country. However, it takes much time to transcribe them into modern Japanese characters since they are difficult to read even for Kuzushiji's experts. As a result, a majority of these books have not yet been transcribed into modern Japanese characters, and most of the knowledge, history, and culture contained within these texts are inaccessible for people.

In 2017, our center (CODH) provided the Kuzushiji dataset to organize the 21st Pattern Recognition and Media Understanding Algorithm Contest for Kuzushiji recognition [2]. The competition focused on recognizing isolated characters or multiple lines of hentaigana, which are easier than whole documents, like in Figure 1. Nguyen et al. won the competition by developing recognition systems based on convolutional neural network and Bidirectional Long Short-Term Memory [3]. In our previous work, we proposed a human-inspired reading system to recognize Kuzushi characters on PRMU Algorithm Contest [4]. The recognition system based on an attention-based encoder-decoder approach to simulate human reading behavior. We achieved better accuracy than the winner of the competition. Tarin et al. presented Kuzushi-MNIST, which contains ten classes of hiragana, Kuzushi-49, which contains 49 classes of hiragana, and Kuzishi-Kanji, which contains 3832 classes of Kanji [5]. The datasets are benchmarks to engage the machine learning community into the world of classical Japanese literature.

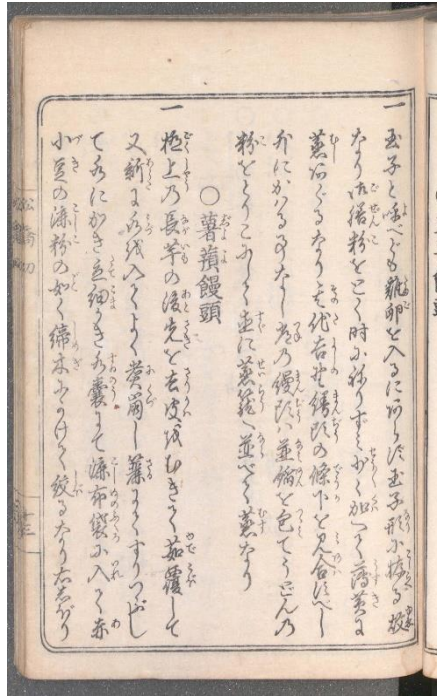

**Fig. 1.** An example of Kuzushi document which contains cursive and connected characters.

The above works for isolated characters and multiple lines of Hetaigana are not suitable for making transcriptions of full-page Kuzushiji documents. Recently, our center organized the Kuzushiji Recognition competition on Kaggle, which requires participants to make predictions of character class and location on documents[6]. The competition provides the ground truth bounding boxes of characters in documents, and the task for participants is to predict character locations and classes. Here, We briefly summarize the top method in the competition.

Tascj employed ensembling models of two Cascade R-CNN for character detector and recognition [7]. The final model achieved 95% accuracy on the private testing set and was ranked the first place in the competition.

Konstantin Lopuhin got second place in the competition [8]. He employed Faster-RCNN model with Resnet backbone for character detection and recognition. For improving the accuracy, he also utilized pseudo labels for the testing set in the training process and ensembling of six models. The final model achieved 95% accuracy on the private testing set but was ranked second place in the competition due to the higher number of submissions.

Kenji employed a two-stage approach and False Positive Predictor [9]. He utilized Faster RCNN for character detection and five classification models for character recogntion. The False Positive Predictor is used for removing false detection. The final model achieved 94.4% accuracy on the private test set and was ranked third place in the competition.

We observed that there are 13 teams achieved accuracy higher than 90%. Most of them employed an object detection based approach, which requires bounding boxes of characters, ensembling models, and data augmentation. However, to make transcription, a process of making transcription from the character classes and locations is needed. The above systems need post-processing to generate text lines. In this paper, we enlarge our previous human-inspired recognition system from multiple lines of Hentaigana to the full-page of Kuzushiji documents. The system can generate transcription from an input image without any post-processing. For the convergence problem of the recognition system for full-page documents, we employ the curriculum learning approach that trains the recognition system step by step from the easy level (several text lines of documents) to the difficult level (full-page documents). For the lack of training data, we propose a random text line erasure approach that randomly erases text lines and dis-torts documents.

The following of this paper is organized as follows. Section 2 briefly describes the overview of the human-inspired recognition system. Section 3 presents the random text line erasure approach for data generation. Section 4 presents the curriculum learning approach for training the recognition system on the full-page Kuzushiji document dataset. Finally, Section 5 and 6 draw the experiments, discussion, and a conclusion of the paper.

## 2   Overview of Human Inspired Recognition System

Our recognition system is based on the attention-based encoder-decoder approach. The architecture of our recognition system is shown in Figure 4. It contains two modules: a DenseNets for feature extraction and an LSTM Decoder with an attention model for generating the target characters. We employed a similar setting for the system as our previous works [4]. The advantage of our model is that it requires images and corresponding transcriptions without bounding boxes of characters.

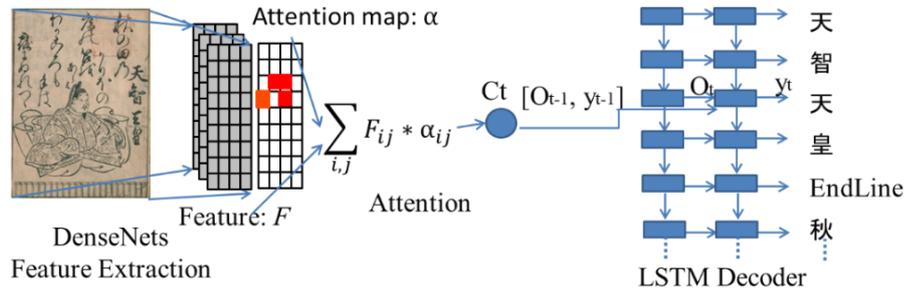

**Fig. 2.** The architecture of the Human Inspired Recognition System.

At each time step $t$, the decoder predicts symbol $y_t$ based on the embedding vector of the previous decoded symbol $E_{y_{t-1}}$, the current hidden state of the decoder $h_t$, and the current context vector $c_t$ as the following equation:

$$p(y_t | y_1, \ldots, y_{t-1}, F) = softmax(W(E_{y_{t-1}} + W_h * h_t + W_c * c_t))$$

The hidden state is calculated by an LSTM. The context vector $c_t$ is computed by the attention mechanism.

## 3   Data Generation

### 3.1   Data Preprocessing

As mentioned earlier, the Center for Open Data in Humanities, the National Institute of Japanese Literature, and the National Institute of Informatics hosted the Kuzushiji recognition on Kaggle. The training dataset provides complete bounding boxes and character codes for all characters on a page. Since our recognition system requires the output as text lines, we have to preprocess the data provided by the competition. We need to preprocess the data to make text lines. We concatenate bounding boxes of characters into vertical lines. Then, we sorted vertical lines from right to left to make the ground truth for the input image. Figure 3 shows an example of the preprocessing process. Note that the competition does not provide bounding boxes for annotation characters (small characters), so the annotation characters should be ignored during the recognition process.

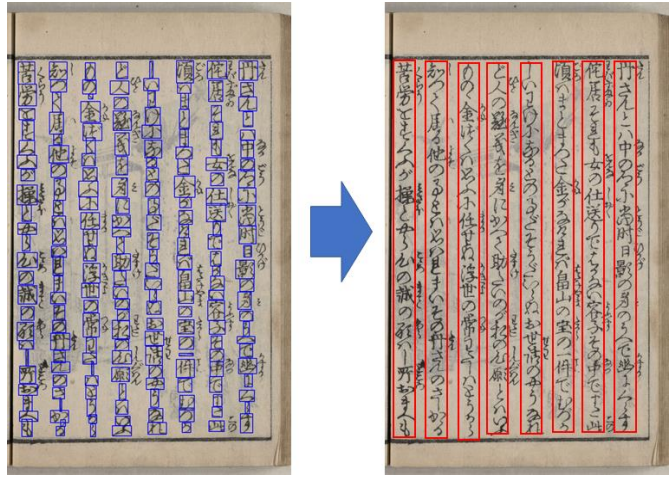

**Fig. 3.** An example of the preprocessing to make text lines from bounding boxes of characters.

### 3.2 Random Text Line Erasure

To train a large deep learning system like Human Inspired Recognition System, we need a massive number of labeled data. However, the dataset for the Kuzushiji document is tiny (around 4000 images for training). Here, we integrate a random text line erasure and perspective skewing to generate more data from the available data. The random text line erasure forces the recognition system to learn to recognize line by line. The perspective skewing transforms the angle of looking images. Here, we employ the left-right skew to preserve the features of vertical text lines. This helps us to improve the performance of the recognition system.

The process of data generation is shown in Figure 4. First, we randomly select $k$ text lines from the input image. Then, we remove the selected text lines from the ground truth and also erase the corresponding text lines in the input image. For erasing text lines, we replace the color of the text lines by the background color of the image. Finally, we employ elastic distortion to create the final image. Figure 5 shows an example of the random text line erase, while Figure 6 shows that of elastic distortion. The red line in Figure 6 is the mask of the perspective skewing. We ensure the content of images does not lose when we use the perspective skewing.

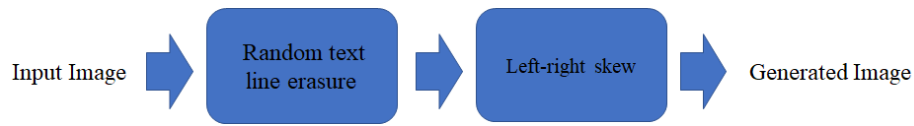

**Fig. 4.** The process of the random text line erasure.

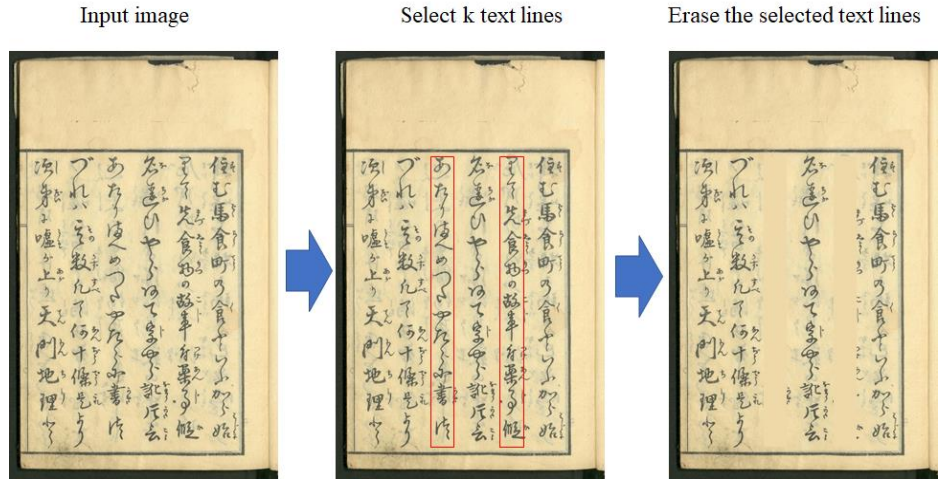

**Fig. 5.** The process of the random text line erasure.

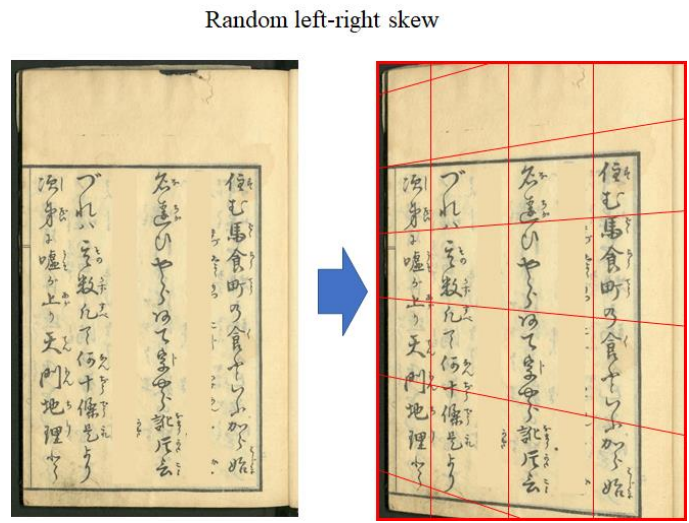

**Fig. 6.** The process of the random skew left-right.

## 4 Curriculum Learning for Human-Inspired Recognition System

We can not train the human-inspired recognition system on full-page of documents because the system is not converged. The first reason is that the problem of recognizing the full page of Kuzushiji documents is very hard. The second reason is the limitation of memory, so that we train the system with a small batch size. Therefore, it affects to speed of learning, the stability of the network during training. To solve this problem, we employ curriculum learning which had proposed by Elman et al. in 1993 [10] and then shown to improve the performance of deep neural networks in several tasks by Bengio.in 2009 [11]. Curriculum learning is a type of learning which starts with easy examples of a task and then gradually increases the task difficulty. Humans have been learning according to this principle for decades. We employ this idea to train the recognition system.

First, we construct an easy dataset by split full-page documents into single or multiple lines. Figure 7 shows the process of generating multiple lines dataset. In the previous work [4], we were able to train the recognition on multiple lines. Recognition of multiple lines is easier than that of full-page documents. Therefore, we first train the recognition with multiple lines images. Then, we add full-page images to the training set. Finally, we add the generated dataset to the training set. Figure 8 shows the three-stage of curriculum learning with different datasets.

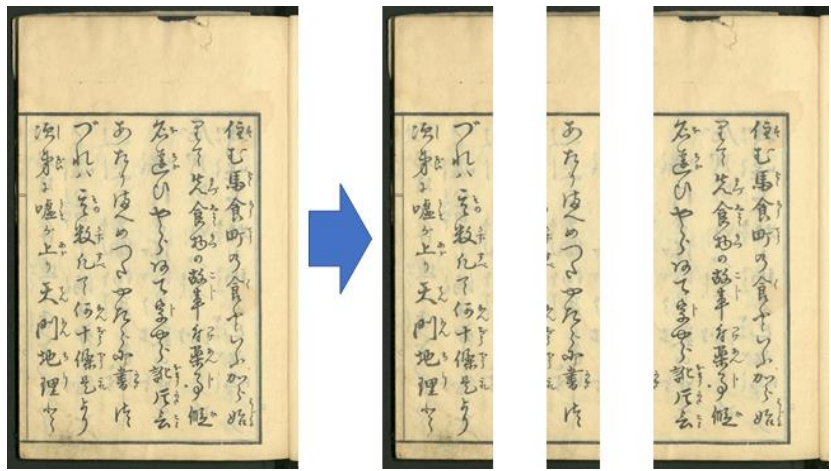

**Fig. 7.** An example of multiple lines generation

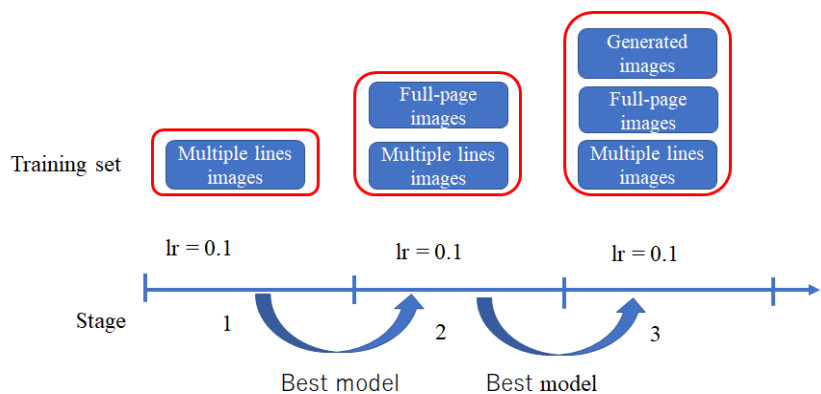

**Fig. 8.** Three stages of the curriculum learning.

## 5  Evaluation

### 5.1  Dataset

We employ the Kuzushiji dataset in the Kaggle competition to train and evaluate our recognition system. The competition provides 3881 images for training and 4150 images for testing. We divide the training images into the origin training and validation set as the ratio 9:1, respectively. As a result, we have 3493 for training and 388 images for validation. For additional training datasets, we employ random lines splitting and data generation to create multiple lines and generated datasets. For testing, we employ the Kaggle public and private testing sets and prepare 400 images with transcription ground truth from Mina de honkoku project [12]. The number of images for training, validation, testing datasets is shown in table 1.

**Table 1. Statistics of the training, validation, and testing datasets**.

| Dataset | # of images |
|---|---|
| Origin training | 3,493 |
| Multiple lines | 9,499 |
| Random text line erasure | 3,493 |
| Validation | 388 |
| Kaggle Public Testing | 1,369 |
| Kaggle Private Testing | 2,781 |
| Transcription Testing | 400 |

### 5.2  Evaluation Metrics

In order to measure the performance of our recognition system, we use two metrics: Character Recognition Rate (CRR) for evaluating transcription generation and F1 for predicting character class and location. Character Recognition Rate is shown in the following equations:

$$CRR(T, h(T)) = 100 - \frac{1}{Z} \sum_{(I,s) \in T} ED(s, h(I))$$

Where $T$ is a testing set which contains input-target pairs ($I$, $s$), $h(I)$ is the output of a recognition system, $Z$ is the total number of target character in $T$ and $ED(s, h(I))$ is the edit distance function which computes the Levenshtein distance between two strings $s$ and $h(I)$.

F1 metrics is generally employed for evaluating character detection task. The detail of the metrics are shown in the following equations:

$$Precision = \frac{Number\ of\ corect\ predicted\ characters}{Number\ of\ predicted\ characters}$$

$$Recall = \frac{Number\ of\ corect\ predicted\ characters}{Number\ of\ groundtruth\ characters}$$

$$F1 = 2 * \frac{Precision * Recall}{Precision + Recall}$$

### 5.3 Training

We train the recognition system in 3 stages as Figure 8. We call the best system in each stage as $S_1$, $S_2$, $S_3$, respectively. For each stage, we used the AdaDelta algorithm with gradient clipping to learn the parameters. The learning rate was set to 0.1. The training process was stopped when the CRR on the validation set did not improve after ten epochs.

### 5.4 Experimental results

In the first experiment, We evaluate $S_1$, $S_2$, $S_3$ on the transcription testing set. Table 2 shows the CRRs of different recognition systems. By employing curriculum learning, we improve the CRR from 86.28% to 88.19%. By data generation, we improve the CRR from 88.19% to 89.51%. If we do not use curriculum learning, the recognition systems are not converged during training. This result verified that curriculum learning and data generation is useful and effective for full-page document recognition.

**Table 2. Performance of the recognition systems on transcription testing set.**

| System | Transcription Testing CRR(%) |
|---|---|
| $S_1$ | 86.28 |
| $S_2$ | 88.19 |
| $S_3$ | 89.51 |

Since our recognition systems predict only transcriptions without the locations of characters, we do not have location information for completing the task in the Kaggle competition. However, as the description in session 2, we use the attention mechanism for predicting a character. Therefore, the attention mechanism may track the location of the character in the input image. We employ the following heuristic to get the location of a character. When the decoder predicts a character, the location of the character is set as the maximum probability of the attention map. We create submission files containing character class and location information for the systems $S_1$, $S_2$, $S_3$ on the second experiment. Table 3 shows the F1 score on the Kaggle public and private testing sets. We achieved 81.8% and 81.9% of F1 score on the Kaggle public and private testing sets, respectively. Although our result is lower than the top participants which achieved 95%, our systems do not use any location information of characters during training. The top participants had used many techniques to improve the accuracy that we have not done yet. For example, they used the ensembling of many models and data augmentation. They used object detection to detect locations of characters while we do not use the locations of characters during training.

Table 3. Performance of the recognition systems on Kaggle public and private testing sets.

| System | Kaggle Public Testing F1(%) | Kaggle Private Testing F1(%) |
|---|---|---|
| $S_1$ | 74.4 | 74.5 |
| $S_2$ | 81.3 | 81.4 |
| $S_3$ | 81.8 | 81.9 |

Since the F1 score on Kaggle competition is low, we visualize the location of character by attention mechanism on several samples to know more about false prediction. Figure 9 shows the recognition result for a part of a document. Blue characters and yellow dots are predicted characters and locations by the recognition system. The bounding boxes are from the ground truth. The correct characters are in red bounding boxes, while incorrect characters are in purple and green bounding boxes. From our observation, there are two types of frequently incorrect predictions. The first type is that the system predicts an incorrect character but a correct location as characters in purple bounding boxes in Figure 9. The second type is that the system predicts a correct character but an incorrect location as characters in green bounding boxes in Figure 9. The second type makes the F1 score in Kaggle testing sets low.

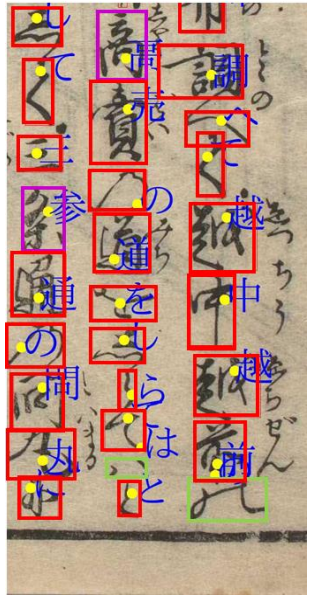

**Fig. 9.** An example of incorrect predicted characters.

Figure 10 shows an example of recognition result in the transcription testing set. Characters with red marks are incorrect recognition. Some of them may be revised by a language model.

Based on the above frequently incorrect prediction, we suggest future works to improve accuracy. For the first type of error, we should make more variety of character shapes such as applying distortion on every character in documents. For the second type of error, we may use location information of characters to supervise the attention model during the training process.

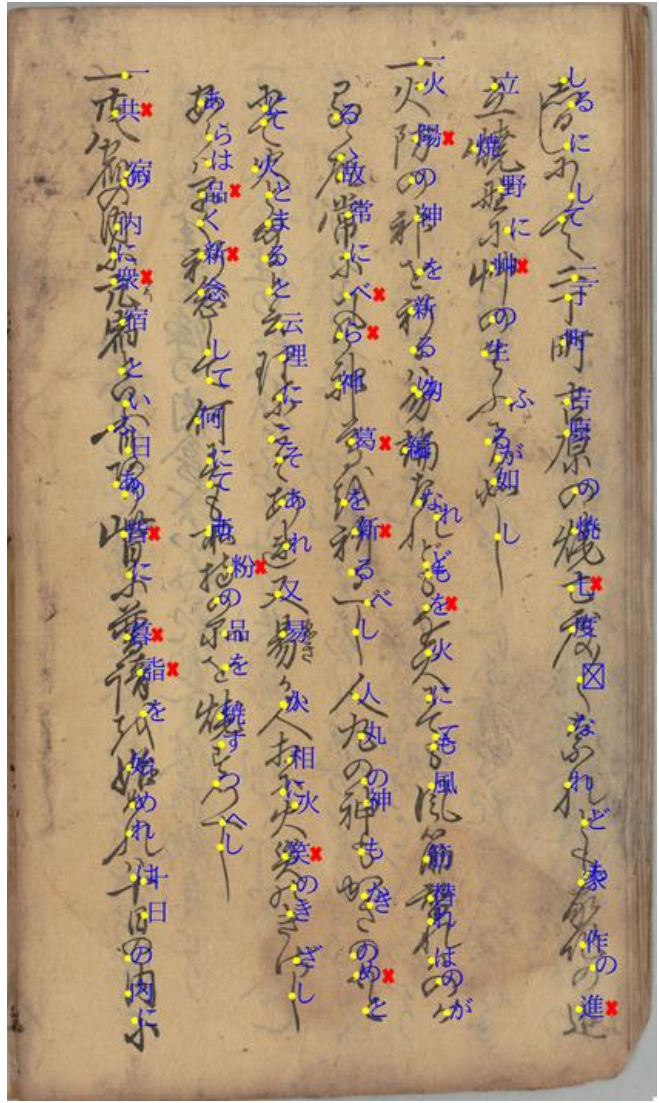

**Fig. 10.** Examples of recognition results on the transcription testing set.

## 6    Conclusion

In this paper, we have proposed the random text line erasure for data generation and training the human-inspired recognition system for full-page of Japanese historical documents by curriculum learning. The efficiency of the proposed system was demonstrated through experiments. We achieved 89.51% of CRR and 81.9% of F1 score on transcription testing and Kaggle testing sets, respectively. We plan to improve the detection system by post-processing in the future.

## References

1. Takahiro, K., Syllabary seen in the textbook of the meiji first year. the bulletin of jissen women's junior college. pages 109–119, 2013
2. Kuzushiji challenge http://codh.rois.ac.jp/kuzushiji-challenge/
3. Hung Tuan Nguyen, Nam Tuan Ly, Cong Kha Nguyen, Cuong Tuan Nguyen, Masaki Nakagawa: Attempts to recognize anomalously deformed Kana in Japanese historical documents, Proc. of the 2017 Workshop on Historical Document and Processing, pp. 31-36, Kyoto, Japan (11.2017).
4. Anh Duc Le, Tarin CLANUWAT, Asanobu KITAMOTO, "A human-inspired recognition system for pre-modern Japanese historical documents", IEEE Access, pp. 1-7, 2019
5. Tarin Clanuwat, Mikel Bober-Irizar, Asanobu Kitamoto, Alex Lamb, Kazuaki Yamamoto, David Ha, Deep Learning for Classical Japanese Literature, arXiv:1812.01718.
6. Kuzushiji Recognition: https://www.kaggle.com/c/kuzushiji-recognition/
7. First place solution: https://www.kaggle.com/c/kuzushiji-recognition/discussion/112788
8. Second place solution: https://www.kaggle.com/c/kuzushiji-recognition/discussion/112712
9. Third place solution: https://www.kaggle.com/c/kuzushiji-recognition/discussion/113049
10. Elman, L., Learning and development in neural networks: the importance of starting small. In: 48, pp. 71–99, 1993.
11. Bengio, Yoshua et al., Curriculum learning. In: Proceedings of the 26th annual international conference on machine learning, pp. 41–48, 2009.
12. Mina de honkoku project: https://honkoku.org/